\documentclass[letterpaper, 10 pt, conference]{ieeeconf}  % Comment this line out if you need a4paper

\IEEEoverridecommandlockouts                              % This command is only needed if 
\usepackage{amsmath} % assumes amsmath package installed
\usepackage{amssymb}  % assumes amsmath package installed
\usepackage{booktabs}
\usepackage{makecell}
\usepackage{multirow}
\usepackage{xcolor}
\usepackage{graphicx}
\usepackage{kotex}
\usepackage{caption}
\usepackage{cleveref}

\title{\LARGE \bf
VG3T: Visual Geometry Grounded Gaussian Transformer
}

\author{Junho Kim$^{1}$, Seongwon Lee$^{1}$$^\dagger$% <-this % stops a space
\thanks{$^{1}$J.\,Kim, S.\,Lee are with the School of Electrical Engineering, Kookmin University, Seoul 02707, South Korea, {\tt\small \{jhk00,sungonce\}@kookmin.ac.kr}}
}

\makeatletter
\let\@oldmaketitle\@maketitle
\renewcommand{\@maketitle}{%
    \@oldmaketitle
    \centering
    \includegraphics[width=\textwidth]{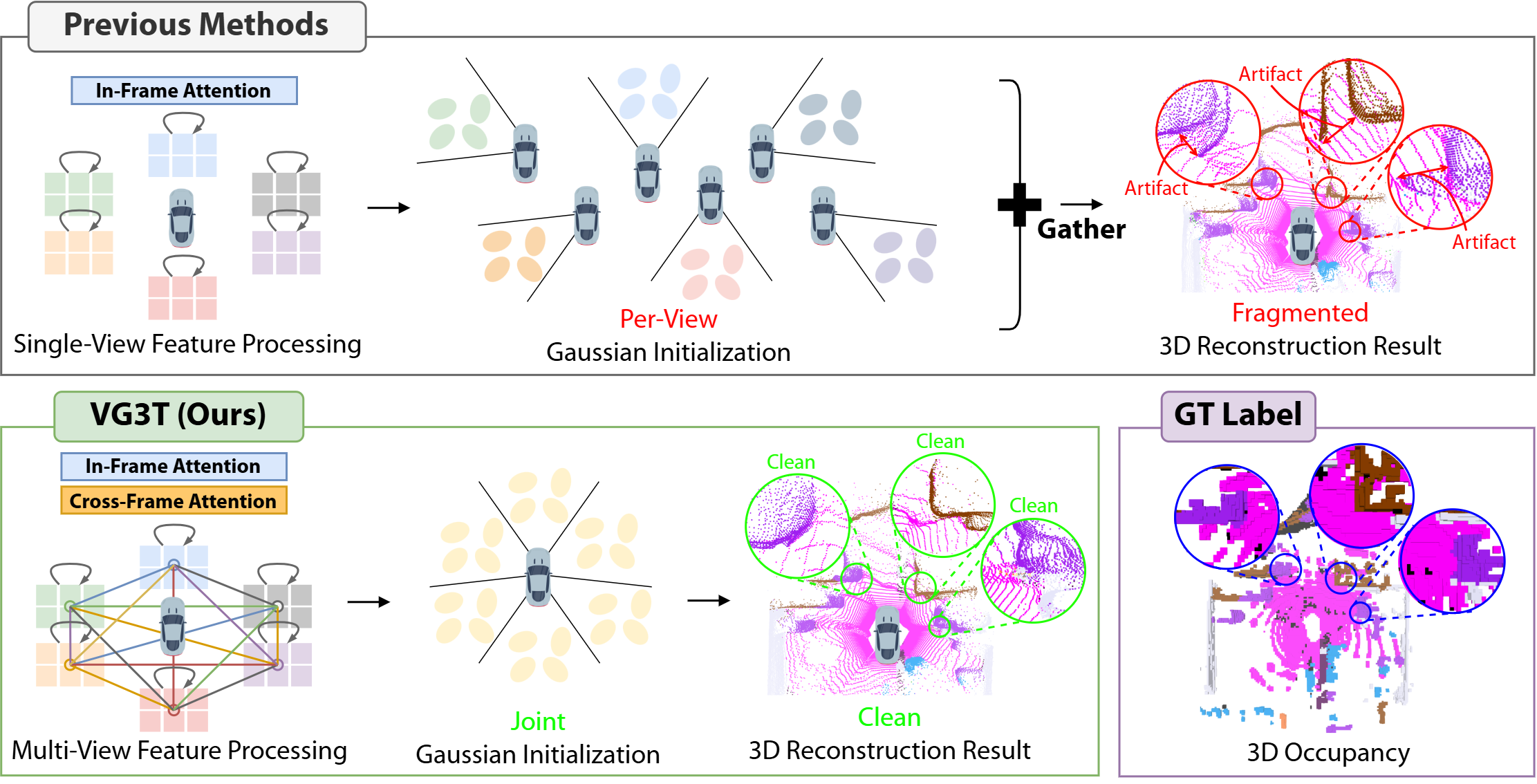}
    \captionof{figure}{\textbf{Coherent 3D Scene Representation via Early Multi-View Fusion.} Unlike prior work that processes each camera view independently, leading to fragmented and inconsistent 3D representations, our method, VG3T, leverages early cross-view correlation. This paradigm produces a coherent and geometrically accurate 3D representation.}
    \label{fig:fig1}
    \vspace{-0.3cm}
    \setcounter{figure}{1}
}
\makeatother

\begin{document}

\newcommand{\red}[1]{{\color{red}#1}}
\newcommand{\todo}[1]{{\color{red}#1}}
\newcommand{\TODO}[1]{\textbf{\color{red}[TODO: #1]}}

\definecolor{nbarrier}{RGB}{255, 120, 50}
\definecolor{nbicycle}{RGB}{255, 192, 203}
\definecolor{nbus}{RGB}{255, 255, 0}
\definecolor{ncar}{RGB}{0, 150, 245}
\definecolor{nconstruct}{RGB}{0, 255, 255}
\definecolor{nmotor}{RGB}{200, 180, 0}
\definecolor{npedestrian}{RGB}{255, 0, 0}
\definecolor{ntraffic}{RGB}{255, 240, 150}
\definecolor{ntrailer}{RGB}{135, 60, 0}
\definecolor{ntruck}{RGB}{160, 32, 240}
\definecolor{ndriveable}{RGB}{255, 0, 255}
\definecolor{nother}{RGB}{139, 137, 137}
\definecolor{nsidewalk}{RGB}{75, 0, 75}
\definecolor{nterrain}{RGB}{150, 240, 80}
\definecolor{nmanmade}{RGB}{213, 213, 213}
\definecolor{nvegetation}{RGB}{0, 175, 0}

\maketitle
\thispagestyle{empty}
\pagestyle{empty}

%%%%%%%%%%%%%%%%%%%%%%%%%%%%%%%%%%%%%%%%%%%%%%%%%%%%%%%%%%%%%%%%%%%%%%%%%%%%%%%%
\begin{abstract}

Generating a coherent 3D scene representation from multi-view images is a fundamental yet challenging task. Existing methods often struggle with multi-view fusion, leading to fragmented 3D representations and sub-optimal performance. To address this, we introduce VG3T, a novel multi-view feed-forward network that predicts a 3D semantic occupancy via a 3D Gaussian representation.
Unlike prior methods that infer Gaussians from single-view images, our model directly predicts a set of semantically attributed Gaussians in a joint, multi-view fashion. This novel approach overcomes the fragmentation and inconsistency inherent in view-by-view processing, offering a unified paradigm to represent both geometry and semantics. We also introduce two key components, Grid-Based Sampling and Positional Refinement, to mitigate the distance-dependent density bias common in pixel-aligned Gaussian initialization methods.
Our VG3T shows a notable 1.7\%p improvement in mIoU while using 46\% fewer primitives than the previous state-of-the-art on the nuScenes benchmark, highlighting its superior efficiency and performance.

\end{abstract}

%%%%%%%%%%%%%%%%%%%%%%%%%%%%%%%%%%%%%%%%%%%%%%%%%%%%%%%%%%%%%%%%%%%%%%%%%%%%%%%%
\section{INTRODUCTION}

Vision-centric systems~\cite{hu2022uniad},~\cite{li2022bevdepthacquisitionreliabledepth},~\cite{chen2017multi},~\cite{philion2020lss} are gaining prominence in autonomous driving due to their inherent cost-effectiveness compared to LiDAR-based solutions~\cite{qi2017pointnetdeephierarchicalfeature},~\cite{qi2017pointnetdeeplearningpoint},~\cite{lang2019pointpillarsfastencodersobject},~\cite{zhou2018voxelnet},~\cite{liang2022bevfusion},~\cite{liu2023bevfusion},~\cite{ye2023lidarmultinet}. However, these systems face a critical challenge: a lack of explicit 3D geometric and semantic understanding. This limitation manifests as fragmented 3D scene representations and can compromise safety when navigating complex and dynamic environments. The emergence of 3D semantic occupancy prediction directly addresses this limitation by generating a fine-grained representation of the surrounding environment, jointly capturing both its geometry and semantic meaning~\cite{cao2022monoscene},~\cite{jiang2023symphonize},~\cite{li2023voxformer},~\cite{li2023fb},~\cite{selfocc},~\cite{occnet},~\cite{wei2023surroundocc},~\cite{zhang2023occformer},~\cite{yu2023flashocc}. While promising, 3D occupancy methods have struggled with efficiency, primarily due to the use of dense representations like voxels. These voxel-based methods, while offering high fidelity, are computationally intensive and inefficient as they fail to leverage the inherent spatial sparsity of real-world scenes. This has motivated a shift towards more efficient and sparse representations~\cite{lu2023octreeocc},~\cite{tang2024sparseocc}. A promising direction models occupancy as 3D Gaussians with learnable attributes~\cite{huang2024gaussian},~\cite{huang2024probabilistic}. This approach is particularly compelling due to its ability to capture the spatial sparsity of real-world scenes, providing a flexible and differentiable representation that models complex geometries with high computational efficiency.

Despite the promise of 3D Gaussian representations, existing methods, such as GaussianFormer~\cite{huang2024gaussian} and GaussianFormer-2~\cite{huang2024probabilistic}, face two critical challenges. First, they rely on a fragmented, view-by-view paradigm where features from each camera are processed independently. This approach makes it difficult to establish coherent cross-view correspondences, resulting in inconsistent and geometrically fragmented 3D representations. A second challenge is a distance-dependent density bias, where Gaussians are over-sampled near the camera and under-represented in distant regions. This imbalance compromises both efficiency and accuracy, as it leads to redundant primitives and a lack of fine-grained detail where it is most needed. These two limitations, fragmented multi-view fusion and density bias, collectively hinder the generation of accurate and efficient 3D occupancy predictions vital for autonomous driving.

In this paper, we introduce Visual Geometry Grounded Gaussian Transformer, dubbed as VG3T, a novel multi-view feed-forward network that directly addresses these challenges. VG3T models a set of semantically-attributed 3D Gaussians from surround-view images. Unlike prior work~\cite{huang2024gaussian},~\cite{huang2024probabilistic}, our model is a unified, end-to-end framework that leverages early multi-view feature correlation. This allows us to fuse multi-view information more coherently, leading to precise geometric and semantic predictions that capture the scene's full 3D structure without relying on external supervisory data. To mitigate the critical distance-dependent density bias, our approach employs a two-staged strategy. First, a Grid-Based Sampling module efficiently removes redundant Gaussians from over-represented areas, guaranteeing an even spatial distribution. Second, the Residual Refinement module precisely adjusts the properties of the remaining Gaussians, enabling them to better capture fine details in previously under-represented areas.

Our main contributions can be summarized as follows:
\begin{itemize}
\item We present VG3T, a novel multi-view feed-forward network that directly predicts 3D semantic occupancy with a Gaussian representation. By leveraging early cross-view feature correlation, our model overcomes the fragmentation inherent in existing view-by-view approaches, enabling a more unified and coherent 3D scene understanding.

\item To address the fundamental problem of distance-dependent density bias, we propose a two-staged strategy: Grid-Based Sampling, which efficiently removes redundant primitives from over-represented areas, and a Positional Refinement module, which precisely adjusts remaining Gaussians to capture fine-grained details.

\item Our end-to-end trainable model, VG3T, achieves a new state-of-the-art on the nuScenes 3D semantic occupancy benchmark. We demonstrate a notable 1.7\%p improvement in mIoU while using 46\% fewer primitives than the previous state-of-the-art method, highlighting the superior efficiency and effectiveness of our approach.

\end{itemize}

\section{Related Work}

\subsection{3D Semantic Occupancy Prediction}
3D semantic occupancy prediction is a critical task for autonomous driving, providing a comprehensive and fine-grained understanding of the surrounding environment. While LiDAR-based methods offer strong performance, their high cost and weather-related vulnerabilities have spurred the development of camera-based alternatives. Early vision-centric methods relied on dense voxel-based representations~\cite{li2023voxformer},~\cite{li2023fb}. While these methods achieve high fidelity, they are computationally and memory-intensive as they inefficiently model empty space. This has motivated the exploration of more efficient scene representations. Another notable direction involves planar-based methods (e.g., BEVFormer~\cite{li2022bevformer}, TPVFormer~\cite{huang2023tri}), which project 3D information onto 2D grids. However, they also suffer from the inefficiency of a dense representation and, more critically, inherently lose crucial spatial information.

This has led to a shift toward more efficient and sparse representations that model only the non-empty parts of a scene. While point-based models offer a sparse representation, each point lacks a defined spatial volume~\cite{shi2024occupancysetpoints},~\cite{wang2024opus}. As a more expressive alternative, 3D Gaussian models (e.g., GaussianFormer~\cite{huang2024gaussian}, GaussianFormer-2~\cite{huang2024probabilistic}) have emerged. These methods capture complex geometries with a compact set of learnable primitives, each defined by attributes like position, shape, and semantics. This representation offers high expressive power while maintaining computational efficiency. However, existing Gaussian methods have limitations. Specifically, they suffer from two critical shortcomings: (1) an inefficient multi-view aggregation scheme that relies on processing each camera view independently, leading to fragmented 3D representations; and (2) a distance-dependent density bias that compromises both accuracy and efficiency. To the best of our knowledge, our proposed VG3T framework is the first to directly and effectively address both of these fundamental challenges, unlocking the full potential of Gaussian representations for 3D occupancy prediction.

\begin{figure*}
  \centering
  % \vspace{-5mm}
  \includegraphics[width=1.0\textwidth]{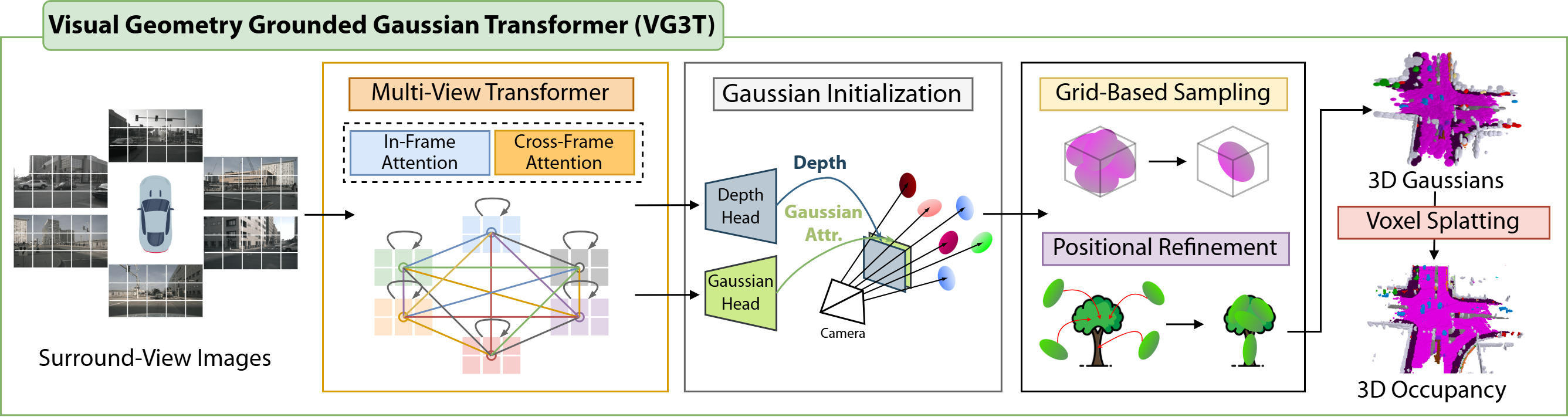}
  \caption{Overview of the Visual Geometry Grounded Gaussian Transformer (VG3T) Architecture.}
  \label{fig:fig2}
  \vspace{-5mm}
\end{figure*}

\subsection{Multi-view 3D Reconstruction}
For 3D semantic occupancy prediction, a critical aspect of multi-view systems is how they effectively aggregate cross-view information. Most contemporary methods employ a late-fusion strategy, where features are first extracted from each camera independently before being fused into a common 3D representation~\cite{li2023voxformer},~\cite{li2022bevformer},~\cite{huang2023tri},~\cite{huang2024gaussian}~\cite{huang2024probabilistic}. The fundamental limitation of this approach is its lack of explicit geometric correspondence at the initial image feature level. This can result in geometric inconsistencies and fragmented 3D representations, a key bottleneck for robust scene understanding.

In contrast, the field of multi-view reconstruction has made significant progress by establishing dense correspondences through the joint processing of multiple images~\cite{wang2024dust3rgeometric3dvision},~\cite{leroy2024groundingimagematching3d},~\cite{Yang_2025_Fast3R}. Recent models, such as Visual Geometry Grounded Transformer (VGGT)~\cite{wang2025vggtvisualgeometrygrounded}, demonstrate this by directly ingesting multiple images to produce a unified geometric understanding. Inspired by this success, our work introduces this early-fusion paradigm to the domain of 3D semantic occupancy prediction. By employing a pre-trained VGGT as our feature backbone, we leverage rich, geometrically-consistent features from the very beginning of our pipeline. This early, coherent fusion provides a robust foundation, enabling our model to overcome the limitations of late-fusion methods and produce a more accurate and consistent 3D representation.

\section{Visual Geometry Grounded Gaussian Transformer (VG3T)}
In this section, we introduce a novel end-to-end multi-view network for 3D semantic occupancy prediction, named Visual Geometry Grounded Gaussian Transformer (VG3T).

\subsection{Problem Setup and Overview}
The goal of 3D semantic occupancy prediction is to generate a comprehensive understanding of the surrounding environment from multi-view images, capturing both its fine-grained geometry and semantic meaning. Specifically, given a set of multi-view input images $\mathcal{I}=\{\mathbf{I}_i\}_{i=1}^{N}$, where ${N}$ is the number of camera views, the task is to predict a dense semantic occupancy grid $\mathbf{O}\in \mathcal{C}^{X\times Y\times Z}$. Here, X, Y, Z define the spatial dimensions of the grid, and $\mathcal{C}$ represents the number of semantic classes.

Unlike traditional methods that directly regress dense voxel grids, VG3T adopts a more efficient and flexible intermediate representation: a set of sparse, learnable 3D Gaussian primitives. Our network is designed to predict a set of 3D Gaussians, $\mathcal{G}=\{\mathbf{G}_i\}_{i=1}^{P}$, where $P$ is the number of Gaussians, which collectively model the geometry and semantics of the scene. Each Gaussian primitive $\mathbf{G}_i$ is a parametric entity defined by a set of learnable attributes: its mean (position) $m_i \in \mathbb{R}^3$, scale $s_i \in \mathbb{R}^3$, rotation $r_i \in \mathbb{R}^4$, opacity $a_i \in [0,1]$, and semantic features $c_i \in \mathbb{R}^C$. The final semantic occupancy grid $\mathbf{O}$ is then rendered by probabilistically aggregating this collection of predicted Gaussians into a dense voxel grid. As shown in Figure \ref{fig:fig2}, our end-to-end framework consists of four main stages: a multi-view feature backbone, an initial Gaussian prediction, a two-staged module comprising Grid-Based Sampling and Positional Refinement, and a final rendering step.

\subsection{Multi-View Transformer Backbone}
The core of VG3T is a powerful feature backbone designed to build a holistic and geometrically consistent understanding of the scene from the earliest stages. Existing methods typically extract features from each camera view independently before a late-fusion step, resulting in fragmented 3D representations. To overcome this fundamental limitation, we leverage the Visual Geometry Grounded Transformer (VGGT)~\cite{wang2025vggtvisualgeometrygrounded} as our backbone. The VGGT architecture, characterized by its minimal 3D inductive biases, enables our model to learn complex geometric correlations directly from large-scale multi-view data.

Specifically, we adapt the VGGT architecture to process features from six surround-view cameras. First, each input image $\mathbf{I}_i\in \mathbb{R}^{3 \times H \times W}$ is tokenized via DINOv2~\cite{oquab2023dinov2} feature extractor into a set of visual tokens $t^I_i\in \mathbb{R}^{K \times C}$, where $K$ is the number of tokens and $C$ is the feature dimension of each token. These tokens are augmented with a small set of learnable register tokens $t^R$, and the full set of tokens from all six views is aggregated into a single sequence. This sequence is then processed by the VGGT's alternating attention mechanism, which strategically interleaves in-frame attention (refining features within each view) and cross-frame attention (fusing features across all views).

The in-frame attention operates on the concatenated tokens of a single view $\mathbf{T}_i=\{{t^I_i,t^R_i}\}$. It refines features based on local semantic and geometric context within each camera view independently. The process is formulated for each view $i\in\{1,…,N\}$ by first computing the query, key, and value matrices via linear projections:
\begin{equation}
\mathbf{Q}_i = \mathbf{T}_i W_{in}^Q, \quad \mathbf{K}_i = \mathbf{T}_i W_{in}^K, \quad \mathbf{V}_i = \mathbf{T}_i W_{in}^V.
\end{equation}

The refined tokens are then computed using the scaled dot-product attention formula:
\begin{equation}
\mathbf{T}_i' = \text{Attention}(\mathbf{Q}_i, \mathbf{K}_i, \mathbf{V}_i) = \text{softmax}\left(\frac{\mathbf{Q}_i\mathbf{K}_i^T}{\sqrt{d_k}}\right)\mathbf{V}i.
\end{equation}

The cross-frame attention then applies self-attention across the entire combined set of tokens from all views, $\mathcal{T} = \bigcup_{i=1}^{N} \mathbf{T}_i'$. This is the crucial step for early multi-view fusion, enabling tokens from any one view to aggregate information from tokens across all other views. This global operation is formulated as:
\begin{equation}
\mathbf{Q} = \mathcal{T}W_{cross}^Q, \quad \mathbf{K} = \mathcal{T}W_{cross}^K, \quad \mathbf{V} = \mathcal{T}W_{cross}^V.
\end{equation}
\begin{equation}
\mathcal{T}' = \text{Attention}(\mathbf{Q}, \mathbf{K}, \mathbf{V}) = \text{softmax}\left(\frac{\mathbf{QK}^T}{\sqrt{d_k}}\right)\mathbf{V}.
\end{equation}

This alternating pattern ensures that each token is iteratively refined by both its in-frame and cross-frame context. This approach is our key to building a robust 3D representation, as it ensures that the visual tokens are enriched with multi-view geometric awareness from the very beginning of the pipeline. The final output is a set of refined visual tokens, $\hat{t}^I_i$, which serve as a powerful foundation for the subsequent dense prediction of 3D Gaussian primitives.

\subsection{Gaussian Initialization}
Following the multi-view feature fusion in the transformer backbone, the cross-view aware visual tokens $\hat{t}^I_i$ for each view are decoded into dense feature maps. Then we use Dense Prediction Transformer (DPT) \cite{ranftl2021visiontransformersdenseprediction} layer to transform the refined output token $\hat{t}^I_i$, into dense feature maps ${\mathbf{F} \in \mathbb{R}^{C \times H \times W}}$. These rich, cross-view correlated feature maps serve as the common input for two parallel prediction heads.

The first head, the Depth head, is responsible for predicting the per-pixel geometry. It consists of a convolutional layer, which we denote as $f_{depth}$. This head processes the dense feature map $\mathbf{F}$ to produce a dense depth map $\mathbf{D} \in \mathbb{R}^{(H/r) \times (W/r)}$ at a down-sampled ratio $r$, along with corresponding aleatoric uncertainty maps $\Sigma^D \in \mathbb{R}+^{(H/r) \times (W/r)}$. This uncertainty is incorporated into our training objective, as it helps the model reason about its confidence in each depth prediction. The process is summarized as follows:

\begin{equation}
    (\mathbf{D}, \Sigma^D) = f_{depth}(\mathbf{F}).
\end{equation}

Concurrently, the second head, the Gaussian head, predicts the remaining physical and semantic attributes of the Gaussians. This head is implemented as a Multi-Layer Perceptron (MLP), which we denote as $f_{attr}$. It maps the each feature $\mathbf{F}_i$ in the dense feature map $\mathbf{F}$ to Gaussian attributes $\mathbf{A}_i=\{s_i,r_i, a_i, c_i\}$. The prediction is formulated as

\begin{equation}
    \mathbf{A}_i = f_{attr}(\mathbf{F}_i).
\end{equation}

Finally, we initialize a complete set of 3D Gaussians, $\mathcal{G}$, by unprojecting each pixel from all $N$ views into 3D space. For each pixel, the 3D mean $\mu$ is computed using the known camera parameters and the predicted depth $d$:

\begin{equation}
    \mu = \mathbf{o} + d \cdot \mathbf{v}.
\end{equation}

Here, o represents the camera's origin, and v is the viewing ray direction for the corresponding pixel. The scale, rotation, opacity, and semantics are directly assigned from the corresponding Gaussian attribute $A_i$, resulting in the final collection of initialized Gaussians set $\mathcal{G}$.

\subsection{Grid-Based Sampling}
To address the distance-dependent density bias inherited from the initialization stage, we propose a two-staged approach that first removes redundant Gaussians and then adjusts their position. The first stage, grid-based sampling, directly targets the problem of primitive redundancy to generate a more uniform spatial distribution.

To achieve this, we first discretize the continuous 3D space by partitioning it into a regular grid. Given the set of initial Gaussians $\mathcal{G}$, we map each mean position $\boldsymbol{\mu}_i \in \mathbb{R}^3$ to a integer grid coordinate $\mathbf{v}_i \in \mathbb{Z}^3$. This is done by scaling the mean position with a pre-defined grid size $s_g$ and taking the floor of the result. To efficiently group all Gaussians that fall within the same voxel, we convert each 3D integer coordinate $\mathbf{v}_i$ into a unique 1D hash key $k_i$. This combined voxelization and hashing process is formulated as
\begin{equation}
    \mathbf{v}_i = \lfloor \boldsymbol{\mu}_i / s_g \rfloor, \quad k_i = H(\mathbf{v}_i).
\end{equation}

This hashing step allows us to group primitives in linear time by sorting the 1D keys, thereby avoiding computationally expensive 3D neighborhood searches and minimizing latency. 

After grouping the Gaussians by their hash keys, we obtain a set of non-empty voxel groups $\mathcal{V}$, where each voxel $v_i$ in $\mathcal{V}$ contains all Gaussians that share the same associated key. Within each occupied voxel containing multiple Gaussians, we then randomly select a single representative Gaussian $\mathcal{G}'_i$ and prune all others. The final, sampled set of Gaussians, $\mathcal{G}_s=\{\mathbf{G}'_i\}_{\forall i}$, is then formed by collecting the single representative from each voxel. This process effectively down-samples over-represented regions, such as those near the camera, while leaving the original sparse representation of distant areas intact, thereby mitigating the initial density bias and improving efficiency.

\begin{table*}[ht]
    \centering
    % \vspace{-5mm}
    \renewcommand{\arraystretch}{1}
    \setlength{\tabcolsep}{2.5pt}
    \caption{\textbf{3D semantic occupancy prediction results on nuScenes benchmark.}}
    \begin{tabular}{l|c c | c c c c c c c c c c c c c c c c}
        \toprule
        Method
        & IoU
        & mIoU
        & \rotatebox{90}{\textcolor{nbarrier}{$\blacksquare$} barrier}
        & \rotatebox{90}{\textcolor{nbicycle}{$\blacksquare$} bicycle}
        & \rotatebox{90}{\textcolor{nbus}{$\blacksquare$} bus}
        & \rotatebox{90}{\textcolor{ncar}{$\blacksquare$} car}
        & \rotatebox{90}{\textcolor{nconstruct}{$\blacksquare$} const. veh.}
        & \rotatebox{90}{\textcolor{nmotor}{$\blacksquare$} motorcycle}
        & \rotatebox{90}{\textcolor{npedestrian}{$\blacksquare$} pedestrian}
        & \rotatebox{90}{\textcolor{ntraffic}{$\blacksquare$} traffic cone}
        & \rotatebox{90}{\textcolor{ntrailer}{$\blacksquare$} trailer}
        & \rotatebox{90}{\textcolor{ntruck}{$\blacksquare$} truck}
        & \rotatebox{90}{\textcolor{ndriveable}{$\blacksquare$} drive. suf.}
        & \rotatebox{90}{\textcolor{nother}{$\blacksquare$} other flat}
        & \rotatebox{90}{\textcolor{nsidewalk}{$\blacksquare$} sidewalk}
        & \rotatebox{90}{\textcolor{nterrain}{$\blacksquare$} terrain}
        & \rotatebox{90}{\textcolor{nmanmade}{$\blacksquare$} manmade}
        & \rotatebox{90}{\textcolor{nvegetation}{$\blacksquare$} vegetation}
        \\
        \midrule
        MonoScene~\cite{cao2022monoscene} & 23.96 & 7.31 & 4.03 &	0.35& 8.00& 8.04&	2.90& 0.28& 1.16&	0.67&	4.01& 4.35&	27.72&	5.20& 15.13&	11.29&	9.03&	14.86 \\
        
        Atlas~\cite{murez2020atlas} & 28.66 & 15.00 & 10.64&	5.68&	19.66& 24.94& 8.90&	8.84&	6.47& 3.28&	10.42&	16.21&	34.86&	15.46&	21.89&	20.95&	11.21&	20.54 \\
        
        BEVFormer~\cite{li2022bevformer} & 30.50 & 16.75 & 14.22 &	6.58 & 23.46 & 28.28& 8.66 &10.77& 6.64& 4.05& 11.20&	17.78 & 37.28 & 18.00 & 22.88 & 22.17 & 13.80 & \underline{22.21}\\
        
        TPVFormer~\cite{huang2023tri} & 11.51 & 11.66 & 16.14&	7.17& 22.63	& 17.13 & 8.83 & 11.39 & 10.46 & 8.23&	9.43 & 17.02 & 8.07 & 13.64 & 13.85 & 10.34 & 4.90 & 7.37\\
        
        TPVFormer*~\cite{huang2023tri}  & {30.86} & 17.10 & 15.96&	 5.31& 23.86	& 27.32 & 9.79 & 8.74 & 7.09 & 5.20& 10.97 & 19.22 & {38.87} & {21.25} & {24.26} & {23.15} & 11.73 & 20.81\\
        
        OccFormer~\cite{zhang2023occformer} & {31.39} & {19.03} & {18.65} & {10.41} & {23.92} & \underline{30.29} & {10.31} & {14.19} & {13.59} & {10.13} & {12.49} & {20.77} & {38.78} & 19.79 & 24.19 & 22.21 & {13.48} & {21.35}\\
        
        SurroundOcc~\cite{wei2023surroundocc} & \underline{31.49} & \underline{20.30}  & \textbf{{20.59}} & {11.68} & \underline{28.06} & \textbf{30.86} & {10.70} & {15.14} & \textbf{14.09} & \textbf{12.06} & \underline{14.38} & \underline{22.26} & 37.29 & \underline{23.70} & {24.49} & {22.77} & \underline{14.89} & {21.86}  \\

        GaussianFormer~\cite{huang2024gaussian} & 29.83 & {19.10} & {19.52} & {11.26} & {26.11} & {29.78} & {10.47} & {13.83} & {12.58} & {8.67} & {12.74} & {21.57} & {39.63} & {23.28} & {24.46} & {22.99} & 9.59 & 19.12 \\
        
        GaussianFormer-2~\cite{huang2024probabilistic} & 30.56 & {20.02} & \underline{20.15} & \underline{12.99} & {27.61} & {30.23} & \underline{11.19} & \underline{15.31} & {12.64} & {9.63} & {13.31} & \underline{22.26} & \underline{39.68} & {23.47} & \underline{25.62} & \underline{23.20} & 12.25 & 20.73 \\
        
        \midrule
        \textbf{Ours} & \textbf{34.06} & \textbf{21.74} & 19.95 & \textbf{13.63} & \textbf{28.69} & 29.52 & \textbf{12.69} & \textbf{16.02} & \underline{13.77} & \underline{10.64} & \textbf{15.75} & \textbf{23.02} & \textbf{41.74} & \textbf{26.26} & \textbf{27.52} & \textbf{26.44} & \textbf{16.51} & \textbf{25.75} \\
        
        \bottomrule
    \multicolumn{19}{l}{* means supervised by dense occupancy annotations
 as opposed to the original LiDAR segmentation labels.} \\
    \multicolumn{19}{l}{The best and second-best performances are represented by \textbf{bold} and \underline{underline} respectively.} \\
    \end{tabular}
    \label{table:nuscenes_results}
    \vspace{-3mm}
\end{table*}

\subsection{Positional Refinement}
While grid-based sampling effectively addresses primitive redundancy, the complementary challenge is to enrich the sparse, under-represented regions of the scene. To this end, the second stage of our approach, positional refinement, performs a learnable process to refine the positions of the sampled primitives.

We treat the sampled set of Gaussians and their corresponding features as a sparse 3D point, which is processed by a refinement network, $f_{r}$. The refinement network consists of 3D sparse convolutional layer, takes as input the features $\mathbf{F}_i$ assigned to each Gaussian $\mathbf{G}_i$ in the sampled Gaussian set $\mathcal{G}_s$ and predicts a set of weights, $\mathbf{w}_i \in \mathbb{R}^K$, for a predefined basis $\mathbf{B} \in  \mathbb{R}^{K \times 3}$. This basis is composed of vectors aligned with the axes, constraining the adjustments.

This allows our model to learn a robust and expressive positional offset, $\Delta\boldsymbol{\mu}_i$, as a linear combination of the basis vectors. The final refined position, $\mu_i'$, is then computed by adding this offset to the original position, $\mu_i$.

\begin{equation}
    \mathbf{w}_i = f{_{r}}(\mathbf{F}_{i}),\quad
    \Delta\boldsymbol{\mu}_i = \mathbf{B}^T\sigma(\mathbf{w}_i),\quad
    \boldsymbol{\mu}'_i = \boldsymbol{\mu}_i + \Delta\boldsymbol{\mu}_i,
\end{equation}
where $\sigma$ denotes the sigmoid function.

This two-stage approach ensures a well-distributed set of primitives that capture essential scene details while significantly reducing the total number of primitives required, yielding the final set of Gaussians, $\hat{\mathcal{G}}$.

\subsection{3D Occupancy Rendering}
The final stage of our methodology is to render the learned set of refined Gaussian primitives, $\hat{\mathcal{G}}$, into the final dense semantic occupancy grid, $\mathbf{O}$. We follow the probabilistic superposition approach from GaussianFormer-2~\cite{huang2024probabilistic}, interpreting each primitive as a local probability distribution over the scene.

First, the total occupancy probability $\alpha(\mathbf{x})$ at any 3D point is calculated by aggregating the influence of all nearby Gaussians. The influence of each primitive is determined by its opacity and a Gaussian kernel, $\phi(\mathbf{x}; \hat{\mathbf{G}}_i)$, which decays with distance from its center. Assuming each contribution is an independent event, the overall probability of occupancy at point $\mathbf{x}$:

\begin{equation}
\alpha(\mathbf{x}) = 1 - \prod_{i=1}^{P}\big(1 - a_i \cdot \phi(\mathbf{x};\hat{\mathbf{G}}_i)\big).
\end{equation}

Next, the semantic prediction for an occupied point is a weighted average of each Gaussian's softmaxed semantic features $\tilde{\mathbf{c}}_i$, scaled by its opacity $a_i$. The weight for each Gaussian is determined by its posterior probability, denoted as $p(\mathbf{x}|\mathbf{G}_i)$, which measures its probabilistic influence at that location. The expected semantic vector $\mathbf{e}(\mathbf{x};\hat{\mathcal{G}})$ is formulated as:
\begin{equation}
\mathbf{e}(\mathbf{x};\hat{\mathcal{G}}) = \frac{\sum_{i=1}^{P}p(\mathbf{x}|\hat{\mathbf{G}}_i)a_i\tilde{\mathbf{c}}_i}{\sum_{j=1}^{P}p(\mathbf{x}|\hat{\mathbf{G}}_j)a_j}.
\end{equation}

The final semantic occupancy vector for a given voxel, $\hat{\mathbf{o}}(\mathbf{x})$, is formed by combining the geometric and semantic predictions. The probability for the empty class is defined as $1-\alpha(\mathbf{x})$, while the probabilities for all semantic classes are the semantic predictions $\mathbf{e}(\mathbf{x};\hat{\mathcal{G}})$ weighted by the total occupancy probability $\alpha(\mathbf{x})$.
\begin{equation}
\hat{\mathbf{o}}(\mathbf{x}) = [1-\alpha(\mathbf{x}); \alpha(\mathbf{x})\cdot\mathbf{e}(\mathbf{x};\hat{\mathcal{G}})].
\end{equation}

\subsection{Training Objective}
Our entire VG3T framework is trained as end-to-end using the total loss, $L_{total}$, which is a weighted sum of two primary components: a loss for the final semantic occupancy prediction, $L_{occ}$, and a loss for supervising the intermediate depth prediction head, $L_{depth}$.
\begin{equation}
    L_{total}=\lambda_{occ}L_{occ}+\lambda_{depth}L_{depth},
\end{equation}
where $\lambda_{occ}$ and $\lambda_{depth}$ are scalar weights that balance the two losses.

\textbf{Occupancy Loss.}
Following common practice in semantic occupancy prediction, our occupancy loss is a combination of a per-voxel cross-entropy loss ($L_{ce}$) and the Lovász-Softmax loss ($L_{lov}$). The cross-entropy loss handles the per-class classification, while the Lovász-Softmax loss is particularly effective for optimizing IoU, a crucial metric for this task. The total occupancy loss is the sum of these two terms:
\begin{equation}
    L_{occ}=L_{ce}+L_{lov}.
\end{equation}

\textbf{Depth Loss.}
To supervise the depth prediction head, we employ an aleatoric uncertainty-aware loss, adapting the formulation used in DUSt3R~\cite{wang2024dust3rgeometric3dvision} and VGGT~\cite{wang2025vggtvisualgeometrygrounded}. This loss is composed of two main terms, both weighted by the predicted per-pixel uncertainty map, $\Sigma_i^D$. The first term measures the direct discrepancy between the predicted and ground-truth depth, while the second is a gradient-matching term that enforces local structural consistency by penalizing differences in the depth gradients. The depth loss is defined as:
$
\mathcal{L}_\text{depth}
=
\sum_{i=1}^N(
\| \Sigma_i^D \odot (\hat{D}_i - D_i) \| + \| \Sigma_i^D \odot ({\nabla} \hat{D}_i - {\nabla} D_i) \|
- \alpha \log \Sigma_i^D)
$. Here, we use the L2-norm to measure the discrepancy, and the final term encourages the model to predict high uncertainty for regions where it is less confident.

\begin{figure*}
  \centering
  % \vspace{-5mm}
  \includegraphics[width=1.0\textwidth]{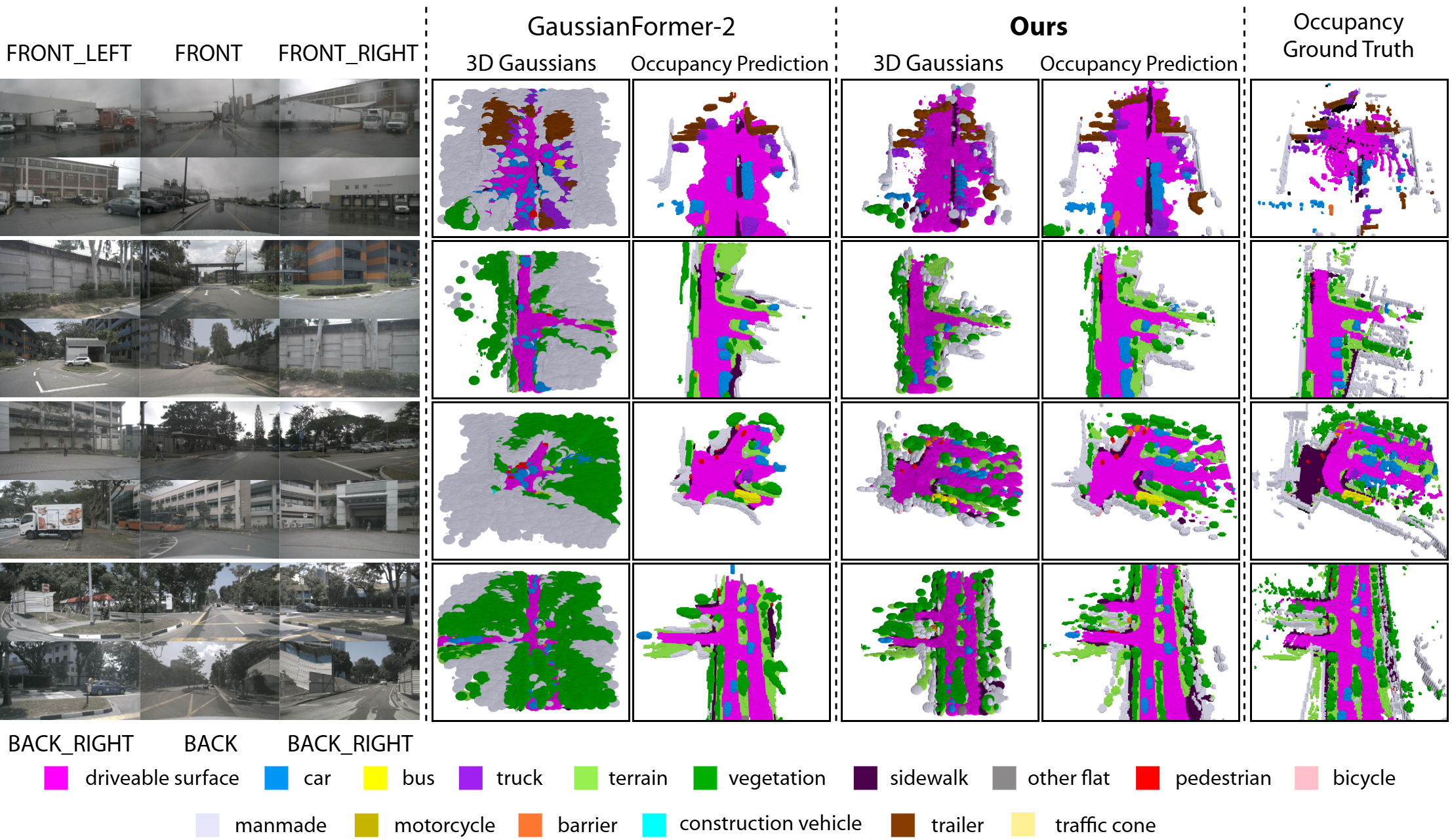}
  \caption{Qualitative results of our method and GaussianFormer-2 on SurroundOcc dataset.}
  \label{fig:fig3}
  \vspace{-3mm}
\end{figure*}

\section{Experiments}

\subsection{Implementation Details}
We implement our VG3T framework using PyTorch. The model's backbone, including the DINOv2~\cite{oquab2023dinov2} patch embedding layers and the main VGGT~\cite{wang2025vggtvisualgeometrygrounded} transformer blocks, is initialized from the official pre-trained VGGT weights and fine-tuned. And all other components, are trained from scratch. The model is trained for 20 epochs on the nuScenes~\cite{caesar2020nuscenes} dataset with a total batch size of 4, distributed across NVIDIA L40S GPUs. For optimization, we use the AdamW optimizer with a weight decay of 0.01~\cite{loshchilov2017adamw}. To ensure stable fine-tuning, a differential learning rate is applied with a cosine annealing schedule: the pre-trained VGGT backbone is optimized with a learning rate of 1e-5, while our newly introduced components are trained with a learning rate of 1e-4. During both training and inference, input images are resized to a resolution of $294 \times 518$. To maintain computational efficiency, the attention operations within the VGGT backbone network are accelerated with FlashAttention-2~\cite{dao2023flashattention2fasterattentionbetter}.

\begin{table}
    \centering
    \caption{Initializing Methods Comparison with GaussianFormer.}
    \setlength{\tabcolsep}{0.01\linewidth}
    \resizebox{1\linewidth}{!}{
    \begin{tabular}{c|c|cc}
        \toprule
        \multirow{2}{*}{Method} & \multirow{2}{*}{Initialization} & \multicolumn{2}{c}{Position}\\
        & & Perc. (\%) \( \uparrow \) & Dist. (m) \( \downarrow \) \\
        \midrule
        GaussianFormer~\cite{huang2024gaussian} & Random & 16.41 & 3.07 \\
        GaussianFormer-2~\cite{huang2024probabilistic} & Single-View & 28.85 & 1.24 \\
        \midrule
        VG3T (ours) & Multi-View & \textbf{51.22} & \textbf{0.97} \\
        \bottomrule
    \end{tabular}}
    \label{table:initializing}
    \vspace{-3mm}
\end{table}

\subsection{Evaluation Benchmarks}
All experiments are conducted on the nuScenes~\cite{caesar2020nuscenes} dataset, a large-scale benchmark for autonomous driving. This dataset consists of 1,000 urban driving sequences, with a standard training, validation, and testing split of 700/150/150. Each 20-second sequence includes imagery from six surround-view cameras, with keyframes annotated at a 2 Hz frequency. We use the dense semantic occupancy annotations from the SurroundOcc~\cite{wei2023surroundocc} benchmark for supervision and evaluation. The ground truth discretizes the scene into a $200 \times 200 \times 16$ voxel grid, covering a spatial volume of $[-50m, 50m]$ on the X/Y axes and $[-5m, 3m]$ on the Z axis. This results in a voxel resolution of $0.5m$. Each voxel is labeled as one of 18 classes: 16 semantic categories, an empty class, and an unknown class.

We evaluate our method using two primary metrics: mean Intersection-over-Union (mIoU) and the standard Intersection-over-Union (IoU). To provide a more comprehensive assessment of scene completion capabilities, we also report Ray-Based IoU (RayIoU)~\cite{tang2024sparseocc} as a supplementary metric. RayIoU specifically evaluates the quality of predictions along rays cast from the camera's viewpoint, offering a robust measure of geometric completion.

\begin{table}
    \centering
    \setlength{\tabcolsep}{0.015\linewidth}
    \scriptsize
    \caption{Efficiency Comparison with GaussianFormer.}
    \scalebox{0.9}{
    \begin{tabular}{l|ccc|ccc}
    \toprule
    Method & \makecell{Number of\\ Gaussians}$\downarrow$ & \makecell{Latency\\ (ms)}$\downarrow$ & \makecell{Memory\\ (MB)}$\downarrow$ & mIoU & IoU & RayIoU\\
    \midrule
    GaussianFormer~\cite{huang2024gaussian} & 144000 & \textbf{372} & 6229 & 19.10 & 29.83 & 25.25 \\ 
    GaussianFormer-2~\cite{huang2024probabilistic} & 25600 & 513 & \textbf{3063} & 20.02 & 30.56 & 29.34 \\
    \midrule
    % FastVG3T (ours) & 13680* & \textbf{342} & 3741 & 21.07 & 33.13 & 31.61 \\
    VG3T (ours) & \textbf{13661*} & 443 & 3746 & \textbf{21.74} & \textbf{34.06} & \textbf{32.66} \\
    \bottomrule
    \multicolumn{7}{l}{* means average number of gaussian.} \\
    \end{tabular}
    }
    \label{table:efficiency}
    \vspace{-3mm}
\end{table}

\begin{figure*}
  \centering
  % \vspace{-5mm}
  \includegraphics[width=1.0\textwidth]{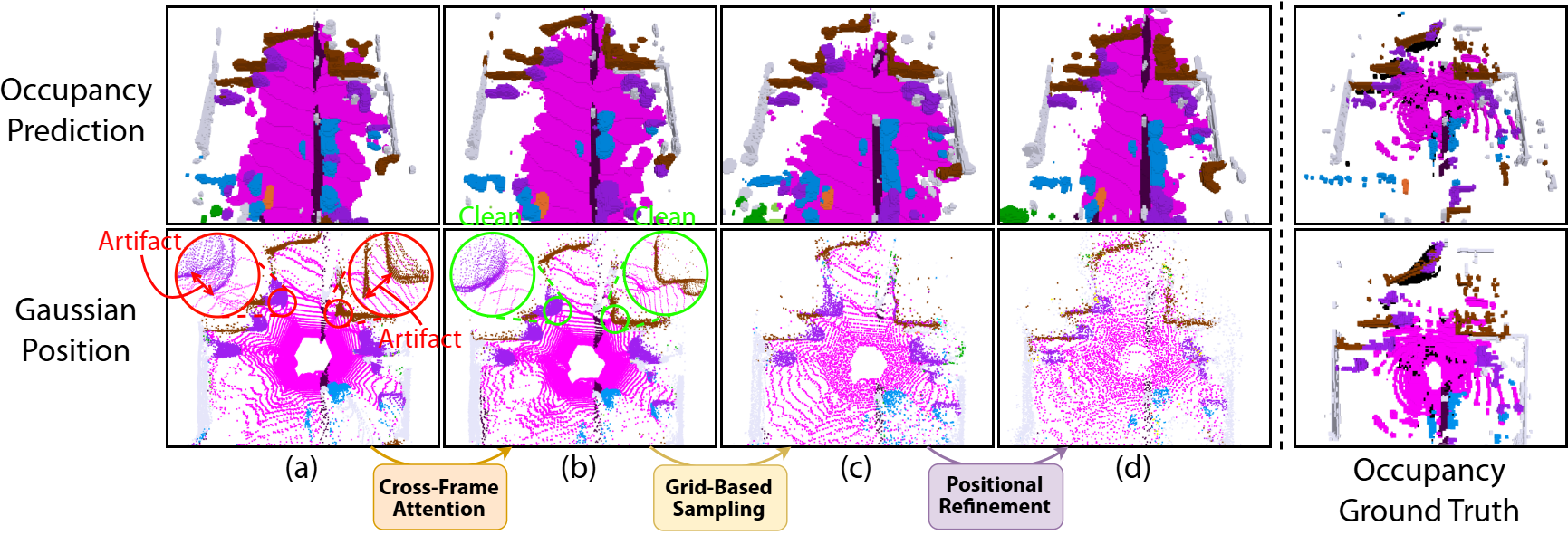}
  \caption{Qualitative results of model component design.}
  \label{fig:fig4}
  \vspace{-3mm}
\end{figure*}

\subsection{Experimental Results}
We evaluate VG3T on the nuScenes~\cite{caesar2020nuscenes} validation set and compare its performance against state-of-the-art methods in (Table \ref{table:nuscenes_results}). Our proposed model sets a new state-of-the-art. This represents a significant improvement of 1.4\%p mIoU over the previous best-performing methods like SurroundOcc~\cite{wei2023surroundocc} and improvement of 1.7\%p over GaussianFormer-2~\cite{huang2024probabilistic}. VG3T demonstrates a clear superiority in modeling static geometry, achieving the highest scores across all background classes. Furthermore, our model shows exceptional performance on small and dynamic object classes, such as bicycle, motorcycle, bus, and trailer, that are traditionally challenging for vision-based methods.

\subsection{Ablation Studies}
We conduct a detailed comparison with the leading Gaussian-based methods in 3D occupancy prediction, GaussianFormer~\cite{huang2024gaussian} and GaussianFormer-2~\cite{huang2024probabilistic}, as shown in Tables \ref{table:initializing} and \ref{table:efficiency}.

\textbf{Initializing Methods Comparison.}
Table \ref{table:initializing} demonstrates the effectiveness of our multi-view initialization strategy. We evaluate initialization quality using two metrics: the percentage of Gaussians positioned in occupied space (Perc.) and the average distance from each Gaussian to its nearest occupied voxel center (Dist.). Our multi-view initialization achieves 77\% improvement in positioning accuracy and 22\% reduction in distance compared to the baseline GaussianFormer-2~\cite{huang2024probabilistic}. This better initialization directly translates to better occupancy prediction performance.

\textbf{Efficiency Comparison.}
Table \ref{table:efficiency} demonstrates that VG3T establishes a new state-of-the-art in both accuracy and efficiency. Compared to the original GaussianFormer, our model achieves a significant 2.64\%p mIoU gain while using 90\% fewer primitives. Furthermore, VG3T surpasses GaussianFormer-2 across all key metrics, achieving a 1.72\%p higher mIoU while being 16\% faster and using 46.6\% fewer Gaussians. This is achieved with a comparable memory footprint, highlighting that our approach breaks the trade-off between performance and speed, delivering superior accuracy with a more efficient representation.

\begin{table}
    \centering
    \setlength{\tabcolsep}{0.008\linewidth}
    \scriptsize
    \caption{Ablation Study on effectiveness of each proposed component.}
    \scalebox{1.0}{
    \begin{tabular}{ccc|cc|ccc}
    \toprule
    \makecell{Multi-View \\ Initialization} & \makecell{Grid\\ Sampling} & Refinement & \makecell{Number of\\ Gaussians}$\downarrow$ & \makecell{Latency\\ (ms)}$\downarrow$ & mIoU & IoU & RayIoU\\
    \midrule
    $\times$ & $\times$ & $\times$ & 64129* & 456 & 21.38 & 33.42 & 32.10 \\
    \checkmark & $\times$ & $\times$ & 64074* & 453 & 21.48 & 33.83 & 32.44 \\
    \checkmark & \checkmark & $\times$ & 14082* & 437 & 21.20 & 33.65 & 31.99 \\
    \checkmark & \checkmark & \checkmark & \textbf{13661}* & \textbf{443} & \textbf{21.74} & \textbf{34.06} & \textbf{32.66}\\
    \bottomrule
    \multicolumn{8}{l}{* means average number of gaussian.} \\
    \end{tabular}
    }
    \label{table:design}
    \vspace{-3mm}
\end{table}

\textbf{Model Component Design.}
We conduct a comprehensive ablation study to validate the effectiveness of our key architectural components, with results summarized in Table \ref{table:design} and visualized in Figure \ref{fig:fig4}. We incrementally add each module to a minimum baseline model to demonstrate its specific contribution to performance and efficiency.

Our minimum baseline, shown in Figure \ref{fig:fig4} (a), processes views independently and operates without any sampling or refinement. Adding Multi-View Initialization establishes an early cross-view correlation, which provides a stronger geometric foundation as seen in (b) and an improvement in both mIoU and RayIoU.

The next stage introduces Grid-Based Sampling, visualized in (c). This module drastically reduces the number of Gaussians by 78\%. This aggressive pruning is crucial for creating a uniform and manageable set of primitives.

The final step, Positional Refinement, addresses this by precisely adjusting the positions of the remaining primitives. As shown in (d), it effectively fills in the details in under-represented areas, recovering the performance lost during sampling. The visual synergy of this two-staged approach is evident in Figure \ref{fig:fig4}, where the final distribution of Gaussians is even and accurately aligned with the scene geometry. Together, these components allow our full model to achieve the best performance with the most efficient representation.

\subsection{Qualitative Results}
We provide qualitative visualizations in Figure \ref{fig:fig3} to illustrate the effectiveness of our approach and compare it against the baseline, GaussianFormer-2~\cite{huang2024probabilistic}. The results demonstrate VG3T's ability to generate comprehensive and geometrically accurate occupancy predictions. GaussianFormer-2’s single-view distribution-based method allocates a significant portion of its primitives to unoccupied space, which reduces its capacity to model fine geometric details accurately, as shown in (Table \ref{table:initializing}). In contrast, our multi-view direct initialization concentrates the model’s entire representational capacity on geometrically relevant regions from the very beginning. Consequently, VG3T achieves superior geometric coverage and detail while leveraging fewer Gaussians than GaussianFormer-2. VG3T is able to capture the sharp, planar surfaces of buildings and other structures with significantly less noise and greater detail compared to the often over-smoothed or incomplete predictions from GaussianFormer-2.

\section{CONCLUSIONS}
This work tackles two fundamental challenges in 3D semantic occupancy prediction: the lack of multi-view correlation from monocular 2D image features leading to fragmented 3D representations, and the density bias inherent in pixel-aligned 3D Gaussian initialization. Our model, VG3T, was designed as a unified solution to both. By employing a multi-view transformer backbone, VG3T effectively exploits cross-view information to initialize all 3D Gaussians in a single, feed-forward pass, enabling end-to-end training. To address the density bias, we introduced a Grid-Based Sampling and Positional Refinement module that manages primitive density. As a result, our proposed model, VG3T, achieves new state-of-the-art performance on the nuScenes benchmark for 3D semantic occupancy prediction.


%%%%%%%%%%%%%%%%%%%%%%%%%%%%%%%%%%%%%%%%%%%%%%%%%%%%%%%%%%%%%%%%%%%%%%%%%%%%%%%%

%%%%%%%%%%%%%%%%%%%%%%%%%%%%%%%%%%%%%%%%%%%%%%%%%%%%%%%%%%%%%%%%%%%%%%%%%%%%%%%%



% \section*{ACKNOWLEDGMENT}

% The preferred spelling of the word ÒacknowledgmentÓ in America is without an ÒeÓ after the ÒgÓ. Avoid the stilted expression, ÒOne of us (R. B. G.) thanks . . .Ó  Instead, try ÒR. B. G. thanksÓ. Put sponsor acknowledgments in the unnumbered footnote on the first page.

%%%%%%%%%%%%%%%%%%%%%%%%%%%%%%%%%%%%%%%%%%%%%%%%%%%%%%%%%%%%%%%%%%%%%%%%%%%%%%%%

\bibliographystyle{IEEEtran}
\bibliography{ref}

\end{document}